\pdfoutput=1

\documentclass[11pt]{article}

\usepackage[]{acl}

\newcommand{\acrofull}[0]{\textbf{K}nowledge \textbf{A}ugmented \textbf{T}ransformer (KAT)}
\usepackage{times}
\usepackage{latexsym}
\usepackage{multirow}
\usepackage{appendix}

\usepackage{times}
\usepackage{latexsym}

\usepackage[T1]{fontenc}

\usepackage[utf8]{inputenc}

\usepackage{microtype}
\usepackage{graphicx}
\usepackage{subfigure}
\usepackage{booktabs}
\usepackage{amsfonts,amssymb}

\usepackage{algorithm}
\usepackage{algorithmic}

\usepackage{amsmath}
\usepackage{booktabs}

\usepackage{newfloat}
\usepackage{listings}
\usepackage{bbold}
	
\usepackage{color, colortbl}
\usepackage{float}
\usepackage{hyperref}
\title{KAT: A Knowledge Augmented Transformer for Vision-and-Language}

\author{
\begin{tabular}{ccc}
        Liangke Gui$^{\mathsection}$$^{\ddagger}$& Borui Wang$^{\dagger}$$^{\ddagger}$ & Qiuyuan Huang$^{\ddagger}$ \\
        Alex Hauptmann$^{\mathsection}$ & Yonatan Bisk$^{\mathsection}$$^{\ddagger}$ & Jianfeng Gao$^{\ddagger}$\\
  {\normalfont $^{\mathsection}$Carnegie Mellon University} & {\normalfont$^{\dagger}$Yale University} & {\normalfont$^{\ddagger}$Microsoft Research} \\
  \multicolumn{3}{c}{\normalfont \texttt{\{liangkeg, alex, ybisk\}@cs.cmu.edu}} \\
    \multicolumn{3}{c}{\normalfont \texttt{borui.wang@yale.edu, \{qihua, jfgao\}@microsoft.com}} \\
\end{tabular}
  }
  
\begin{document}
\maketitle
\begin{abstract}


The primary focus of recent work with large-scale transformers has been on optimizing the amount of information packed into the model's parameters.  In this work, we ask a complementary question: Can multimodal transformers leverage explicit knowledge in their reasoning? 
Existing, primarily unimodal, methods have explored approaches under the paradigm of knowledge retrieval followed by answer prediction, but leave open questions about the quality and relevance of the retrieved knowledge used, and how the reasoning processes over implicit and explicit knowledge should be integrated. To address these challenges, we propose a - \acrofull{} - which achieves a strong state-of-the-art result (+6\% absolute) on the open-domain multimodal task of OK-VQA. Our approach integrates implicit and explicit knowledge in an encoder-decoder architecture, while still jointly reasoning over both knowledge sources during answer generation. 
Additionally, explicit knowledge integration improves interpretability of model predictions in our analysis. Code and pre-trained models are released at \url{https://github.com/guilk/KAT}.
\end{abstract}

\section{Introduction}
\label{sec:intro}

\let\thefootnote\relax\footnote{Work done when Liangke and Borui interned at Microsoft Research.} 
There has been a revival of interest in knowledge-intensive tasks which require an external knowledge source for humans to perform. Many applications in real-world scenarios, such as autonomous AI agents, need to seamlessly integrate implicit (\emph{i.e.}, commonsense) and explicit knowledge (\emph{e.g.}, Wikidata) to answer questions. In this work, we investigate how to effectively integrate implicit and explicit knowledge for reasoning. Tasks like Outside Knowledge Visual Question Answering (OK-VQA)~\citep{marino2019ok} require that models use knowledge not present in the input to answer questions, making it an ideal test bed for investigating this implicit-explicit knowledge trade-off. 

\begin{figure}[tb]
\centering
\includegraphics[width=0.48\textwidth]{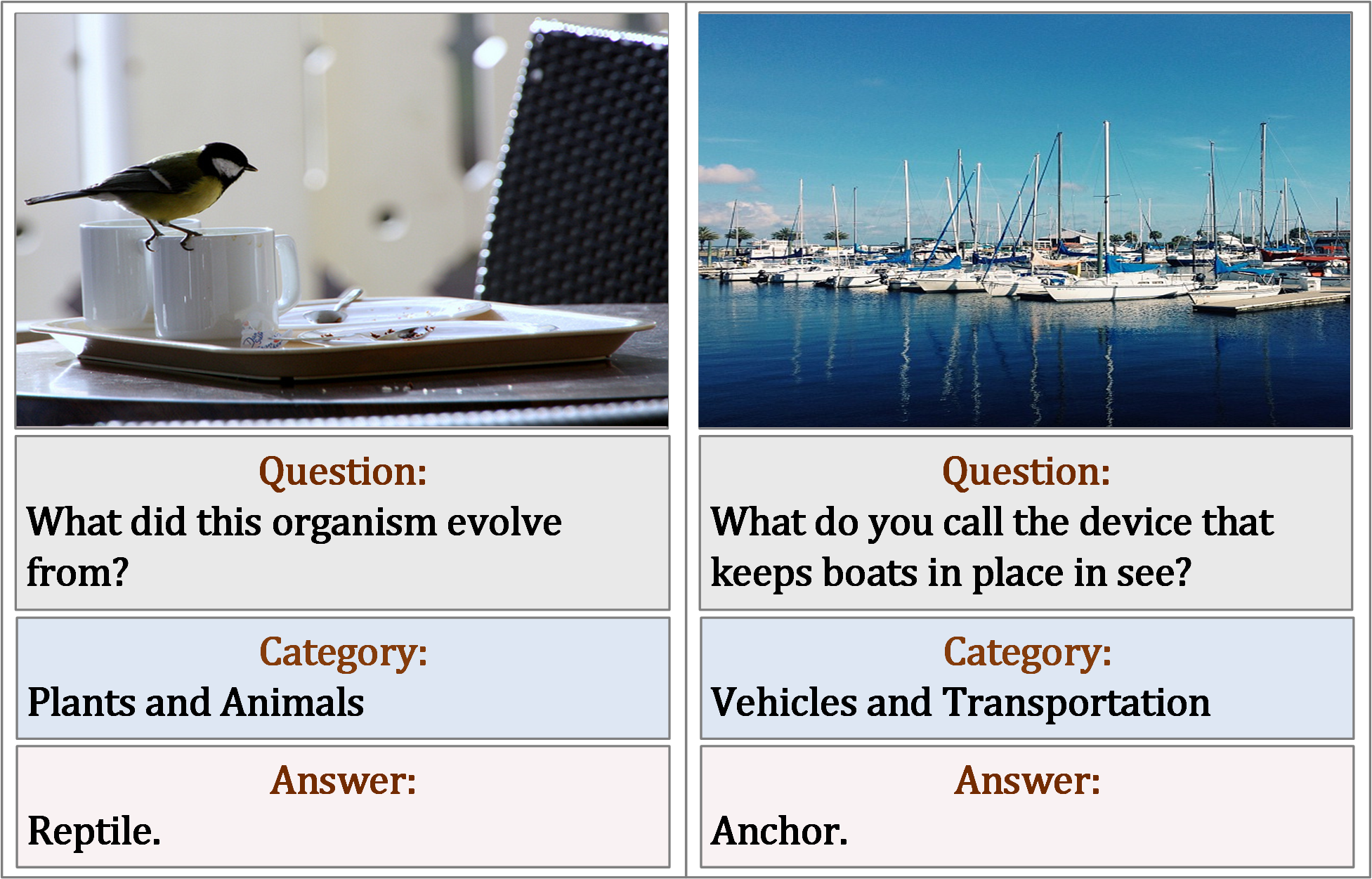}
\caption{Examples of knowledge-based VQA that requires external knowledge. Success on this task requires not only visual recognition, but also logical reasoning to incorporate external knowledge about the world.}
\label{fig:okvqa-demo}
\vspace{-0.3cm}
\end{figure}

Consider the examples from OK-VQA shown in Figure~\ref{fig:okvqa-demo}. To answer the question in the left example, the system needs to both ground \textit{organism} to bird through explicit knowledge and then apply the implicit knowledge \emph{birds evolved from reptiles} to answer the question. Similarly for the question in the right example, the system needs to recognize boats and harbor and requires the implicit knowledge \emph{anchors are used to stop boats from moving}. A key challenge here is to accurately link image content to abstract external knowledge. There have been a number of recent developments demonstrating the feasibility of incorporating external knowledge into Question Answering models~\citep{wang2015explicit, li2020boosting,marino2021krisp, wu2021multi,garderes2020conceptbert}. Existing methods first retrieve external knowledge from external knowledge resources, such as DBPedia~\citep{auer2007dbpedia} and ConceptNet~\citep{liu2004conceptnet} before jointly reasoning over the retrieved knowledge and image content to predict an answer. 

However, most existing approaches have several drawbacks. First, explicit knowledge retrieved using keywords from questions or image tags may be too generic, which leads noise or irrelevant knowledge during knowledge reasoning. Second, existing work mainly focuses on explicit knowledge which is often in the form of encyclopedia articles or knowledge graphs. While this type of knowledge can be useful, it is insufficient to answer many knowledge-based questions. As shown in Figure~\ref{fig:okvqa-demo}, questions require the system to jointly reason over explicit and implicit knowledge, which is analogous to the way humans do.
To address these challenges, we propose an approach, \textbf{KAT}, to effectively integrate implicit and explicit knowledge during reasoning. 
The main contributions of our work are as follows:


\noindent \textbf{i) Knowledge extraction.} We adopt two novel methods for knowledge extraction that significantly improve the quality and relevance of extracted knowledge: for implicit knowledge, we design new prompts to extract both tentative answers and supporting evidence from a frozen GPT-3 model; for explicit knowledge, we design a contrastive-learning-based explicit knowledge retriever using the CLIP model, where all the retrieved knowledge are centered around visually-aligned entities.

\noindent \textbf{ii) Reasoning in an encoder-decoder transformer.} We design a novel reasoning module in \emph{KAT} to perform jointly reasoning over explicit and implicit knowledge during answer generation, which is trained by using an end-to-end encoder-decoder transformer architecture. 

\noindent \textbf{iii) OK-VQA performance.} \emph{KAT} sets a new state of the art on the challenging OK-VQA~\citep{marino2019ok} benchmark, and significantly outperforms existing approaches.


\vspace{0.2cm}

\section{Related Work}
\label{sec:relatedwork}

\paragraph{Vision-Language Transformer.} Multimodal transformers have made significant progress over the past few years, by pre-trained on large-scale image and text pairs, then finetuned on downstream tasks. VisualBERT~\citep{li2019visualbert}, Unicoder-VL~\citep{li2019unicoder}, NICE~\citep{NICE}, and VL-BERT~\citep{su2019vl} propose the single-stream architecture to work on both images and text. ViLBERT~\citep{lu2019vilbert} and LXMERT~\citep{tan2019lxmert} propose a two-stream architecture to process images and text independently and fused by a third transformer in ta later stage. While these models have shown to store in-depth cross-modal knowledge and achieved impressive performance on knowledge-based VQA~\citep{marino2021krisp, wu2021multi,luo2021weakly}, this type of implicitly learned knowledge is not sufficient to answer many knowledge-based questions~\citep{marino2021krisp}. Another line of work for multimodal transformers, such as CLIP~\citep{radford2021learning} or ALIGN~\citep{jia2021scaling}, aligns visual and language representations by contrastive learning. These models achieve state-of-the-art performance on image-text retrieval tasks. Different from existing work that uses multimodal transformers as implicit knowledge bases, we focus primarily on how to associate images with external knowledge. Importantly, our model only relies on multimodal transformers learned by contrastive learning which do not require any labeled images. This makes our model more flexible in real-world scenarios.

\paragraph{Knowledge-based VQA.} Some Knowledge-based visual language tasks requires external knowledge beyond the image to answer a question. Early exploration, such as FVQA~\citep{wang2017fvqa}, creates a fact-based VQA dataset by selecting a fact (\emph{e.g., <Cat, CapableOf, ClimbingTrees>}) from a fixed knowledge base. A recent Outside Knowledge VQA (OK-VQA) dataset is a more challenging dataset, covering a wide range of knowledge categories. In our work, we focus on OK-VQA due to its large-scale knowledge-based questions as well as its open-ended nature.

Recent approaches have shown a great potential to incorporate external knowledge for knowledge-based VQA. Several methods explore aggregating the external knowledge either in the form of structured knowledge graphs~\citep{garderes2020conceptbert, narasimhan2018out,li2020boosting, wang2017fvqa, wang2015explicit}, unstructured knowledge bases~\citep{marino2021krisp,wu2021multi,luo2021weakly}, and neural-symbolic inference based knowledge ~\citep{K-Inference, Symbolic-knowledge}. In these methods, object detectors~\citep{ren2015faster} and scene classifiers~\citep{he2016deep} are used to associate images with external knowledge. Further, external APIs, such as Google~\citep{wu2021multi,luo2021weakly}, Microsoft~\citep{KB-VLP, yang2021empirical} and OCR~\citep{luo2021weakly, wu2021multi} are used to enrich the associated knowledge. Finally, pre-trained transformer-based language models~\citep{KB-VLP, yang2021empirical}, or multimodal models~\citep{wu2021multi, luo2021weakly, wu2021multi, garderes2020conceptbert, marino2021krisp} are leveraged as implicit knowledge bases for answer predictions. 

Different from previous approaches, Our work aims to develop a single, unified architecture, by jointly reasoning over explicit and implicit knowledge to augment generative language models. While part of our approach is similar to PICa~\citep{yang2021empirical} which considers GPT-3 as implicit knowledge base, our model takes one step further by showing that how explicit and implicit knowledge can be integrated during knowledge reasoning. Another similar work Vis-DPR~\citep{luo2021weakly} collects a knowledge corpus from training set by Google Search which is specific to a certain dataset. Our proposed model is more generic by collecting entities from Wikidata and not limited to the training set. 

\paragraph{Open-Domain Question Answering (ODQA).} ODQA is the NLP task of answering general domain questions, in which the evidence is not given as input to the system. Several approaches~\citep{chen2017reading, karpukhin-etal-2020-dense} propose to predict the answers by first retrieving support document from Wikipedia, before extracting answers from the retrieved document. Recent works~\cite{izacard2020leveraging,lewis2020retrieval} combine text retrieval models with language generative models which achieve state-of-the-art performance on knowledge-intensive natural language processing tasks. Similar to these works as part of our method, we extend this framework to VQA domain and show the effectiveness of aggregating explicit and implicit knowledge for knowledge-based VQA.

\section{Method}
\label{sec:method}

\begin{figure*}[t]
    \centering
    \includegraphics[ 
    width=0.92\textwidth]{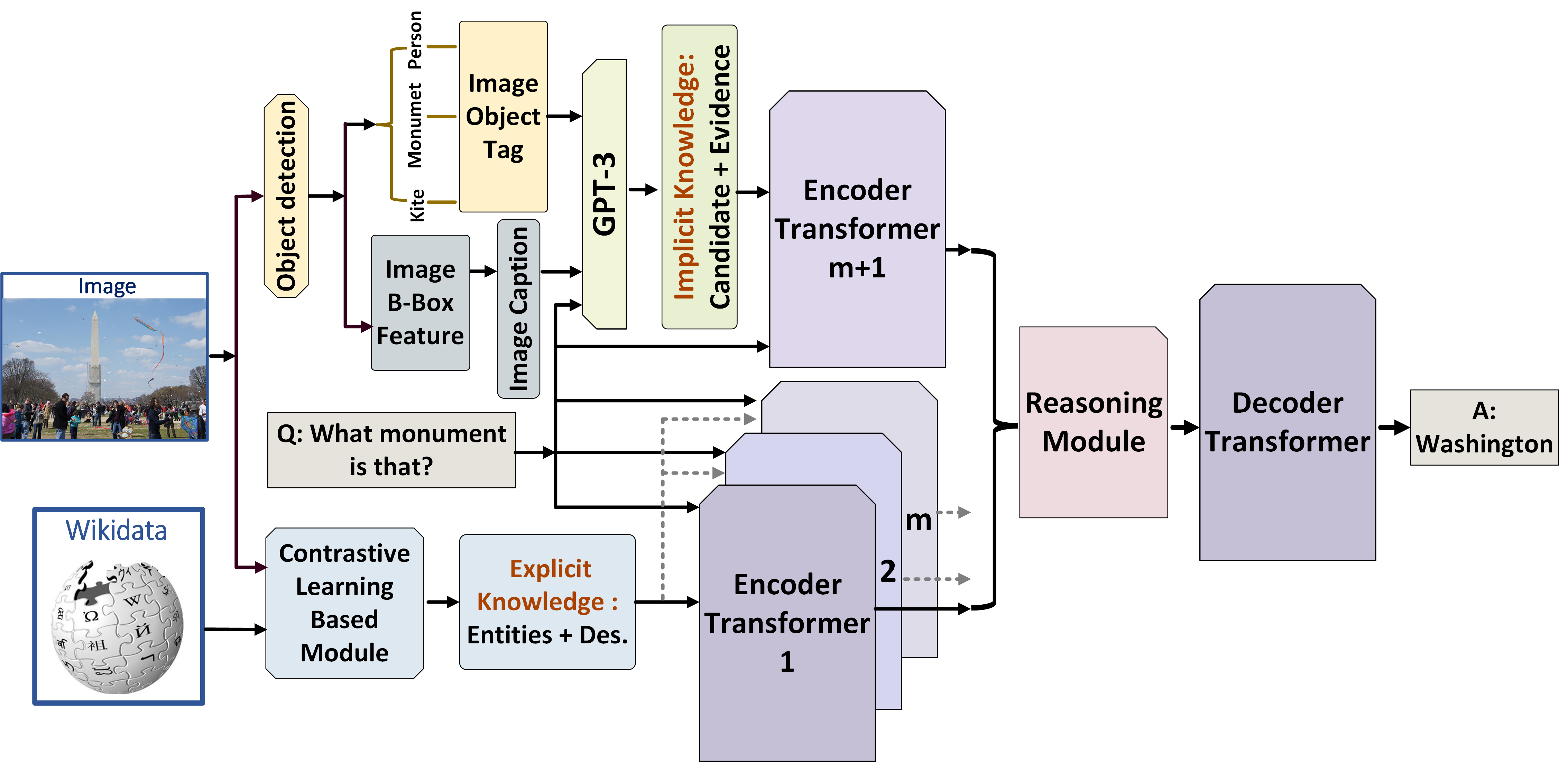}
    \vspace{-2mm}
\caption{Our \emph{KAT} model uses a contrastive-learning-based module to retrieve knowledge entries from an explicit knowledge base, and uses GPT-3 to retrieve implicit knowledge with supporting evidence. The integration of knowledge is processed by the respective encoder transformer, and jointly with reasoning module and the decoder transformer as an end-to-end training with the answer generation.} 
\label{fig:framework}
\vspace{-3mm}
\end{figure*}

\subsection{Overview}
When humans reason about the world, they process multiple modalities and combine external and internal knowledge related to these inputs. Inspired by this idea, we introduce a new \emph{KAT} approach.
The overview of the proposed \emph{KAT} model is shown in Figure~\ref{fig:framework}. 
We define the knowledge from explicit knowledge bases as the explicit knowledge, and the knowledge stored in large-scale language models as the implicit knowledge (\emph{i.e.}, implicit commonsense knowledge). We describe the retrieval method of our explicit knowledge (\S \ref{sec:explicit}) and the retrieval method of our implicit knowledge (\S \ref{sec:implicit}). Next, we introduce the details of our knowledge reasoning module which jointly reasons over both explicit and implicit knowledge (\S \ref{sec:encoder}). 

\paragraph{Problem Formulation.} We apply our \emph{KAT} on OK-VQA task in this paper. Formally, given a training dataset $\mathbb{D} = \{(v_{i},q_{i},a_{i})\}_{i=1}^{s}$, where $v_{i}$ denotes the $i^{th}$ training image; $s$ is the total number of the training images; $q_{i}$ and $a_{i}$ represent the $i^{th}$ question and its corresponding answer, respectively. We use a sequence-to-sequence model that is composed of an encoder and a decoder, which is a comparison method of T5~\cite{2020t5} or BART~\cite{lewis2019bart}. Let $\theta$ be the parameters of the model $p$ that needs to be trained. Unlike previous approaches that treat this task as a classification problem~\cite{wu2021multi, marino2021krisp}, our model is to take $v_i$ and $q_i$ as inputs and generate the answer $a_i$ in an auto-regressive manner. It should be noted that our proposed model tackles a more challenging problem. As the generated answer may contain an arbitrary number of words from the entire vocabulary.

\subsection{Explicit Knowledge Retrieval}
\label{sec:explicit}
\subsubsection{Explicit Knowledge Extraction}
Given an image $v_i$ and corresponding question $q_i$, it is important to ground image regions with fine-grained descriptions, which is conducive to understanding both the image content and the question with the referred items. Existing approaches~\cite{radford2021learning, jia2021scaling} on OK-VQA apply object detectors to generate image tags which are used for explicit knowledge retrieval. Such image tags can be generic and have a limited vocabulary size, leading noise or irrelevant knowledge. Motivated by the recent progress of visual-semantic matching approaches~\cite{radford2021learning, jia2021scaling}, we leverage a contrastive-learning-based model to associate image regions with external knowledge bases.

Similar to the previous work~\citep{marino2021krisp,luo2021weakly} which uses a subset of external knowledge, we construct an explicit knowledge base that covers the 8 categories of animals, vehicles and other common objects from Wikidata~\citep{wikidata}. The details can be found in Section~\ref{sec:kb}. We denote the constructed knowledge base as $\mathcal{K}$. Each knowledge entry $e$ from $\mathcal{K}$ is a concatenation of the entity and its corresponding description. 

The goal of our explicit knowledge retriever is to index all knowledge entries in $d_r$-dimensional dense representations by a dense encoder $E_{ent}(\cdot)$, such that it can efficiently retrieve the top $m$ knowledge entries relevant to each input image. Given an image $v_i$, we use a sliding window with a stride to generate $N$ image regions $\{v_i^1,...,v_i^N\}$. Then an image encoder $E_{img}(\cdot)$ is applied to map each patch to a $d_r$-dimensional dense representation, and retrieves $k$ knowledge entries from $\mathcal{K}$ whose representations are closest to the patch-level representation. To define the similarity score between the image region $v_i^j$ and the entity $e$, we use the inner product of their normalized representations:
\begin{equation}
    sim(v_i^j, e) = E_{ent}(e)^TE_{img}(v_i^j).
\end{equation}
In total, we retrieve the top $N\times k$ knowledge entries relevant to image $v_i$. We keep top-$m$ knowledge entries ranked by similarity scores as explicit knowledge source $x^{exp}$. 

In principle, the image and knowledge entry encoders can be implemented by any multimodal transformer. We use the CLIP model (ViT-B/16 variant)~\cite{radford2021learning} in our work and take the \texttt{[CLS]} as representations. 
We pre-extract representations of the knowledge entries in the knowledge base $\mathcal{K}$ using the entity encoder $E_{ent}$ and index them using FAISS~\cite{johnson2019billion}. 
The qualitative example for the extracting explicit knowledge model is presented in Appendix~\ref{app:explicit}.

\subsubsection{Knowledge Base Construction}
\label{sec:kb}
We use the English Wikidata~\citep{wikidata} dump from Sep. 20, 2021 as the explicit knowledge source base which contains $95,870,584$ entities. Each data item is stored in a structured format constituted of property-value pairs. Properties are objects and have their own Wikidata pages with labels, aliases, and descriptions. We extract a subset that covers common objects in real-world scenarios. We remove all entities whose string labels or corresponding descriptions are empty or non-English. This results in a total of $423,520$ entity triplets in the end (\emph{e.g.}, \emph{<Q2813, Coca-Cola, carbonated brown colored soft drink>}) (See Table~\ref{tab:ontology}).

\begin{table}[!htb]
\centering
\begin{tabular}{ll|c}
Subclass &             & Number\\
\hline
Role & (Q214339) & 162,027 \\
Point of interest & (Q960648) & 85,900 \\
Tool & (Q39546)            & 78,621 \\
Vehicle & (Q42889) & 44,274 \\
Animal & (Q729) & 18,581 \\
Clothing & (Q11460) & 17,711 \\
Company & (Q891723) & 12,173 \\
Sport & (Q349) & 4,233 \\ \hline
Total & & 423,520 \\ \hline
\end{tabular}
\caption{We collect a subset of Wikidata that covers common objects in real-life scenarios as our explicit knowledge base. Above are statistics of these subclasses.}
\label{tab:ontology}
\vspace{-5mm}
\end{table}


\subsection{Implicit Knowledge Retrieval}
\label{sec:implicit}
While our explicit knowledge retriever focuses on semantic matching between image regions and knowledge entries, it lacks implicit commonsense knowledge (\emph{e.g.}, \textit{Lemons are sour}) which is usually stored in large-scale language models~\cite{brown2020language}. In this section, we retrieve implicit knowledge with supporting evidence by prompting from a large-scale pre-trained language model. 

We design our implicit knowledge retriever with inspirations from the previous work~\citep{yang2021empirical}. We leverage GPT-3 as an implicit language knowledge base and treat VQA as an open-ended text generation task. For each image-question pair, we first convert the image $v_i$ into a textual description $C$ via a state-of-the-art image captioning model~\citep{li2020oscar}, and then construct a carefully designed text prompt consisting of a general instruction sentence, the textual description $C$, the question, and a set of context-question-answer triplets taken from the training dataset that are semantically most similar to the current image-question pair (see Figure~\ref{fig:prompt2} in Appendix~\ref{app:prompt} for a concrete example). We then input this text prompt to the GPT-3 model in its frozen version and obtain the output from GPT-3 as the tentative answer candidate to the current image-question pair.

To gain deeper insights from the implicit knowledge coming out of GPT-3 and its rationale, we design another prompt to query GPT-3 for supporting evidence behind the tentative answer candidate that it generates. 
More specifically, for each image-question pair $(v_i, q_i)$, and for a tentative answer $a$ generated by GPT-3, we construct the prompt in the form of:  ``(question $q_i$)? (answer $a$). This is because'' to query GPT-3 for supporting evidence (see Figure~\ref{fig:prompt1} in Appendix~\ref{app:prompt} for a concrete example). We finally compile both the tentative answers and the corresponding supporting evidence from GPT-3 as implicit knowledge source $x^{imp}$.

\subsection{KAT Model}
\label{sec:encoder}

As showed in the Figure~\ref{fig:framework}, the explicit knowledge entries are from an image, which are concerned with semantic matching of the image regions. These knowledge entries could be noisy or irrelevant to its corresponding question. Moreover, some of the supporting evidence prompted from GPT-3 is generic or not related to image content. Simple concatenation of different knowledge may introduce noise during model training. We design a knowledge reasoning module with inspirations from the previous work~\cite{karpukhin-etal-2020-dense}. Our knowledge reasoning module encodes each question and knowledge pair separately, and jointly reason over both explicit and implicit knowledge when generating an answer. 

\paragraph{Encoder.} We concatenate question $q_i$ with each knowledge as a question-knowledge pair. Firstly, we add sentinel tokens \texttt{question:}, \texttt{entity:} and \texttt{description:} before the question, the retrieved entity, and its description separately. Similarly, we add sentinel tokens \texttt{question:}, \texttt{candidate:} and \texttt{evidence:} before the question, the tentative answer, and its evidence. 
Secondly, we use an embedding layer followed by a sequence of encoder layers to encode the question-knowledge pairs separately. We average the token embeddings of each question-knowledge pair from the last encoder layer, which results in an embedding matrix of explicit knowledge $X^{exp} \in \mathbb{R}^{m\times d}$ and implicit knowledge $X^{imp} \in \mathbb{R}^{p\times d}$, where $d$, $m$ and $p$ are the embedding dimension, the number of explicit knowledge $x^{exp}$, and the number of implicit knowledge $x^{imp}$, respectively. 

\paragraph{Reasoning Module.} To jointly reason  over implicit and explicit knowledge, we concatenate the embeddings of explicit and implicit knowledge form a global representation $X \in \mathbb{R}^{(m+p)\times d}$. The cross-attention module takes the global representation $X$ of the encoder as the input. Let $H\in\mathbb{R}^d$ be the output of the previous self-attention layer of the decoder. By definition~\citep{vaswani2017attention}, the scaled dot-product attention can be expressed as: 
\begin{equation}
    Q_v = softmax(\frac{QK^T}{\sqrt{d}})V,
\end{equation}
where queries $Q$, keys $K$, and values $V$ are computed by applying linear transformations: $Q=W_QH, K=W_KX, V=W_VX$. The attended representation $Q_v$ is a weighted sum of the values, and implies that our model performs a joint reasoning over explicit and implicit knowledge when generating answers.

\paragraph{Decoder.} We feed the embeddings of explicit and implicit knowledge to a sequence of decoder layers for answer generation. We train our model with a cross-entropy loss:

\begin{equation}
    \mathcal{L}_{CE} = -\sum_{t=1}^{n} \log p_{\theta}(y_t|y_{<t},x^{exp};x^{imp}),
\end{equation}
where $y_t$ is predicted autoregressively.

\section{Experiment}
\label{sec:experiment}


\begin{table*}[!htb]
\centering
\begin{tabular}{l  l | c | c }
\toprule
& Method & Knowledge Resources & Acc (\%)\\  
\midrule
\multirow{5}{*}{\rotatebox{90}{No knowledge}} & Q only~\citep{marino2019ok} & - &14.93 \\
 & Vanilla T5 & - & 18.56 \\ 
 & MLP~\citep{marino2019ok} & - &20.67\\ 
 & BAN~\citep{marino2019ok} & - &25.1\phantom{0}\\
 & MUTAN~\citep{marino2019ok} & -&26.41\\ 
\midrule
\midrule
\multirow{7}{*}{\rotatebox{90}{With knowledge}}& BAN+AN~\citep{marino2019ok}& Wikipedia &25.61\\
& BAN+KG-AUG~\citep{li2020boosting} & Wikipedia+ConceptNet & 26.71 \\
& MUTAN+AN~\citep{marino2019ok} & Wikipedia &27.84\\
& ConceptBERT~\citep{garderes2020conceptbert} & ConceptNet &33.66\\
& KRISP~\citep{marino2021krisp} & Wikipedia+ConceptNet& 38.35 \\ 
& Vis-DPR~\citep{luo2021weakly} & Google Search & 39.2\phantom{0} \\ 
& MAVEx~\citep{wu2021multi} & Wikipedia+ConceptNet+Google Images& 39.4\phantom{0} \\ \midrule \midrule
\multirow{2}{*}{\rotatebox{90}{\small GPT-3}} & PICa-Base~\citep{yang2021empirical} & Frozen GPT-3 (175B) & 43.3\phantom{0} \\
& PICa-Full~\citep{yang2021empirical} & Frozen GPT-3 (175B) & 48.0\phantom{0} \\ \midrule \midrule
\multirow{4}{*}{\rotatebox{90}{}}& KAT-explicit (w/ reasoning) & Wikidata & 44.25 \\
&  KAT-implicit (w/ reasoning) & Frozen GPT-3 (175B) & 49.72 \\
&  KAT (w/o reasoning) & Wikidata+Frozen GPT-3 (175B) & 51.97 \\
&  KAT (single) & Wikidata+Frozen GPT-3 (175B) & 53.09\\
&  \bf KAT (ensemble) & Wikidata+Frozen GPT-3 (175B) & \bf 54.41\\
\bottomrule
\end{tabular}
\caption{Results of OK-VQA comparing to standard baselines show that our KAT (large size) model achieves state-of-the-art performance on OK-VQA full testing set. It is important (see table sections) to compare methods based on their access to increasingly large implicit sources of knowledge and utilization of explicit knowledge sources. Our five KAT models variants make the relative importance of these decisions explicit. We train our model with $3$ random seeds and the result is denoted as \emph{ensemble}.}
\label{exp:main}
\vspace{-3ex}
\end{table*}

\subsection{Dataset}
\label{sec:dataset}

\noindent\textbf{OK-VQA}~\citep{marino2019ok} is currently the largest knowledge-based VQA dataset, The questions are crowdsourced from Amazon Mechanical Turkers and require outside knowledge beyond the images in order to be answered correctly. The dataset contains $14,031$ images and $14,055$ questions covering a variety of knowledge categories. We follow the standard evaluation metric recommended by the VQA challenge~\citep{antol2015vqa}.

\subsection{Implementation Details} 
\label{sec:implement}
For the knowledge reasoning module, we initialize our model with the pre-trained T5 model~\citep{2020t5}. We compare two model sizes, base and large, each containing $220M$ and $770M$ parameters respectively. We fine-tune the models on OK-VQA dataset, using AdamW~\citep{loshchilov2017decoupled}. We use a learning rate of $3e-5$ to warm up for $2K$ iterations and train for $10K$ iterations. Limited by the computational resources, we set the number of retrieved entities to $40$. The model is trained with a batch size of 32, using 16 V100 GPUs with 32Gb of memory each. Unless otherwise specified, all results reported in this paper as KAT use this model which we found to perform best. We evaluate our predictions with ground-truth after normalization. The normalization step consists of lowercasing, and removing articles, punctuation and duplicated whitespace~\citep{chen2017reading,lee2019latent}. To be consistent with previous work~\cite{marino2021krisp}, we train our model with $3$ different random seeds and use the average results for the leaderboard submission.

\subsection{Comparison with Existing Approaches}
\label{sec:sota}

We compare our model against existing approaches on the OK-VQA dataset and the results are summarized in Table~\ref{exp:main}. Our model outperforms state-of-the-art methods by significant margins. We compare our model with existing approaches from two aspects. (1) If we only consider using explicit knowledge, our model achieves $44.25\%$ which is $4.85\%$ and  $5.9\%$ higher than MAVEx and KRISP, respectively. Our model uses contrastive-learning-based model to extract knowledge, leaving headroom by incorporating supervised pre-trained models, such as pre-trained object detectors. It should be noted that our proposed model is working on a more challenging problem. As the generated answer could contain an arbitrary number of words from the entire vocabulary. Our model is slightly better than PICa-Base which is a plain version of PICa-Full without example engineering. It implies that our single, unified architecture can effectively associate images with the explicit knowledge base. (2) If we take the implicit knowledge from GPT-3 as the additional input, our model outperforms PICa-Full by $6.41\%$ which indicates it is important to integrate knowledge of different types when generating answers. The detailed comparison can be found in Table~\ref{exp:okvqa-ana}.

\definecolor{Gray}{gray}{0.9}
\section{Ablation Study}
\label{sec:ablation}
To unpack the performance gain and understand the impact of different components, we ablate and compare different model architectures, types of knowledge and the number of explicit knowledge. 
\begin{table}[!htb]
\centering
\begin{small}
\resizebox{0.48\textwidth}{!}{\begin{tabular}{c c | c  c | c }
\toprule
 \multicolumn{2}{c|}{\bf Model architecture}  & \multicolumn{2}{c|}{\bf Knowledge}  & \bf Accuracy (\%) \\
Base   & Large  & Explicit & Implicit & \\ 
 \midrule
  $\surd$   &   & &  &  18.56 \\ 
 \rowcolor{Gray}
  $\surd$   &   & $\surd$ &  &  40.93 \\
     & $\surd$  & $\surd$  &  &  44.25\\
 \rowcolor{Gray}
   $\surd$  &   &  & $\surd$ &  47.60 \\
      & $\surd$  &  & $\surd$ &  49.72 \\
  \rowcolor{Gray}
  $\surd$ &  & $\surd$ & $\surd$ & 50.58 \\
  & $\surd$ & $\surd$ & $\surd$  & 54.41 \\
\bottomrule
\end{tabular}}
\caption{Ablation study on model architectures and types of knowledge. Our experiments show that larger model has more capacity for implicit knowledge reasoning and jointly reasoning over both knowledge sources has a consistent improvement with baselines.}
\label{exp:okvqa-ana}
\end{small}
\vspace{-3mm}
\end{table}

Specifically, as shown in Table~\ref{exp:okvqa-ana}, our KAT-large shows a consistent improvement over using KAT-base. This larger model has more capacity for implicit knowledge reasoning. The integration of explicit and implicit knowledge achieves a performance gain of $\sim\! 4\%$, supporting the intuition that these two types of knowledge 
provide complementary pieces of knowledge.

\subsection{Effectiveness of Knowledge Reasoning}

\begin{table}[!htb]
\centering
\begin{tabular}{ l c }
\toprule
Method  & Accuracy (\%)\\  
\midrule
KAT (w/o reasoning) & 51.97 \\
KAT & \bf 54.41\\
\bottomrule
\end{tabular}
\caption{Comparison with KAT (w/o reasoning) which uses the concatenated knowledge as inputs without the knowledge reasoning module.}
\label{exp:krm}
\vspace{-3mm}
\end{table}

To verify the effectiveness of our knowledge reasoning module, we use a KAT without the knowledge reasoning module which is denoted as KAT (w/o reasoning). This model concatenates explicit and implicit knowledge as a sentence and adopts a maximum length of 256 tokens. We train this variant with the same parameter settings. As shown in Table~\ref{exp:krm}, simply concatenating knowledge sources is $2.43\%$ lower than our proposed model. It indicates that KAT (w/o reasoning) may introduce noise to relevant knowledge during encoding. Our model adaptively attend different knowledge sources for answer generation that can reduce the influence of irrelevant knowledge.

\subsection{Extracting Explicit Knowledge}
\begin{figure}[!ht]
    \centering
    \includegraphics[width=0.5\textwidth,clip, trim=20 5 30 35]{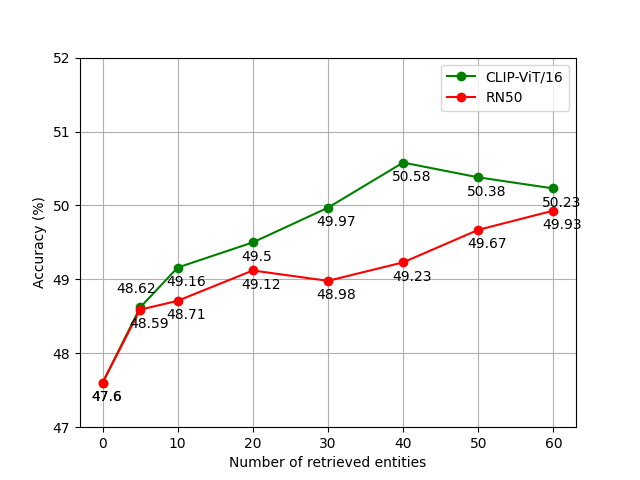}
    \vspace{-2mm}
\caption{Our model achieves consistent improvement when aggregating more knowledge entries from an explicit knowledge base. However, as CLIP-ViT/16 and RN50 are very different explicit knowledge retrieval backbones we see the choice of backbone and number of sources to include are intimately related.   Here we use KAT-base for demonstration.}
\label{fig:numberofentities}
\vspace{-3mm}
\end{figure}

\begin{figure*}[!ht]
    \centering
    \includegraphics[trim={0cm 6cm 0 0},clip, 
    width=0.9\textwidth]{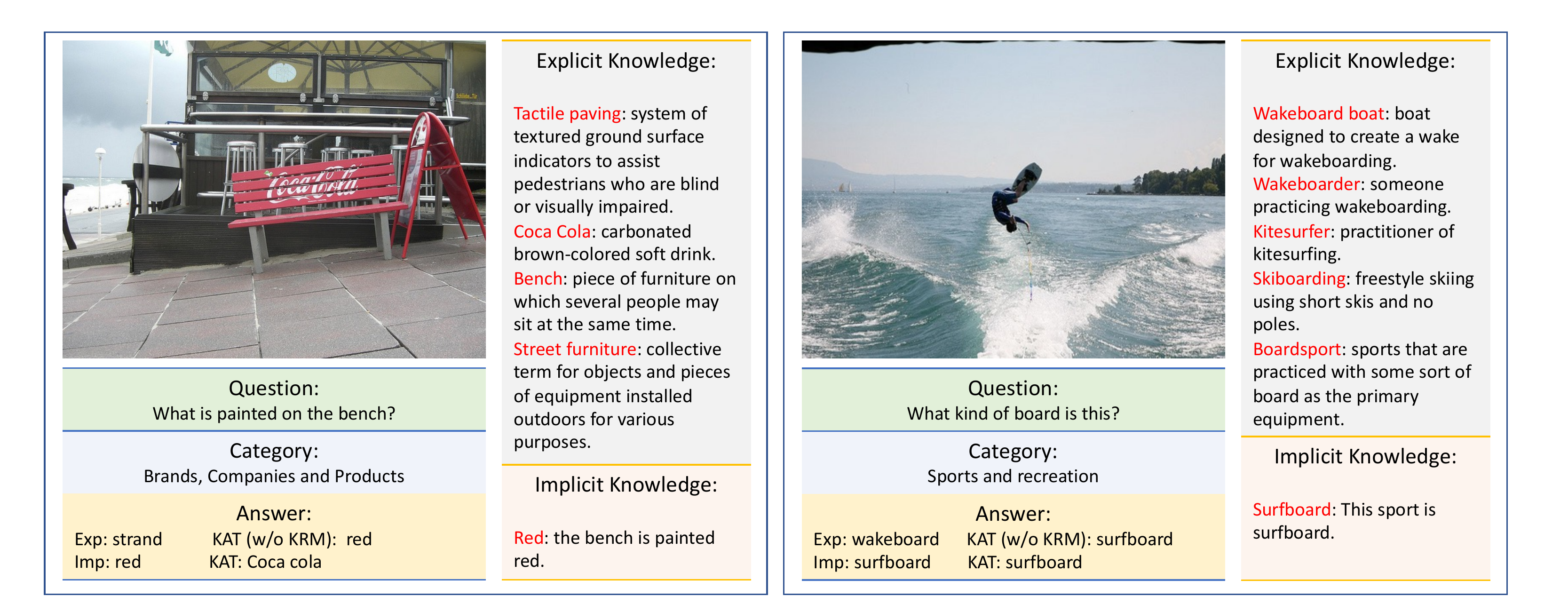}
    \vspace{-2mm}
\caption{Two examples from OK-VQA dataset that our model generates correct answers by jointly reasoning over both implicit and explicit knowledge. (exp: predictions by using explicit knowledge only and imp: predictions by using implicit knowledge only). More examples and analysis can be found in Appendix~\ref{app:analysis}.}
\label{fig:examples}
\vspace{-4mm}
\end{figure*}

From Figure~\ref{fig:numberofentities} we can see, the performance of our model is directly affected by the size of retrieved explicit knowledge. When only considering the implicit knowledge (\emph{i.e.}, the number of retrieved entities is $0$), our model achieves $47.6\%$ which is slightly worse than PICa-Full baseline. It indicates that solely increasing model complexity cannot improve the performance. This also demonstrates the importance of explicit knowledge.
Our model shows a consistent improvement by incorporating more explicit knowledge. While a more extensive knowledge set may include more distracting knowledge, retrieved knowledge entries can share either visually or semantically similar knowledge as the relevant ones. Thus this can massively reduce the search space and/or reduce spurious ambiguity.

We compare different explicit knowledge retrieval module. Though ViT/16 has a large classification improvement over ResNet-50 (\emph{e.g.}, $6.9\%$ on ImageNet)~\citep{radford2021learning}, there is a less gap between these two backbones. As the number of retrieved entities increases, our knowledge reasoning module can further migrate this gap by adaptively attending to different explicit knowledge.

\subsection{Category Results on OK-VQA}
\label{sec:category}
Here we present quantitative analyses to illustrate how explicit and implicit knowledge influence the final predictions. Based on the types of knowledge required, questions in OK-VQA are categorized into 11 categories and the accuracy results of each category are reported in Table~\ref{exp:category}. We re-train our model under the same settings with only either explicit or implicit knowledge, denoted as ``exp'' and ``imp'' respectively. 

For most categories, the model using only explicit knowledge performs worse than that using only implicit knowledge. As implicit knowledge comes from the results of state-of-the-art object detection, image captioning models and supporting evidence by prompting GPT-3. While explicit knowledge is retrieved based on semantic matching between images and entities from knowledge bases, it contains richer but more distracting knowledge. Note that using explicit knowledge performs better for category ``Brands, Companies, and Products" and ``Weather and Climate". It indicates that accurately recognizing objects with fine-grained descriptions in the images is important for these categories to answer corresponding questions.

\begin{table}[!htb]
\centering
\resizebox{0.47\textwidth}{!}{\begin{tabular}{@{}l@{\hspace{5pt}}c@{\hspace{5pt}}c@{\hspace{5pt}}c@{\hspace{5pt}}c@{}}
\toprule
\bf Question Type & Exp  & Imp   &\bf Ours  & \bf $\Delta$ \\
\midrule
Plants and Animals                & 42.2  & 51.5 & 54.7 & \phantom{0}+3.2\\
Science and Technology            & 44.4  & 43.3 & 52.8 & \phantom{0}+8.3 \\
Sports and Recreation             & 49.7  & 53.8 & 60.4 & \phantom{0}+6.7\\
Geo, History, Lang, and Culture   & 45.6  & 45.4 & 55.8 & +10.2\\
Brands, Companies, and Products   & 41.7  & 38.2 & 48.5 & \phantom{0}+6.8\\
Vehicles and Transportation       & 41.5  & 42.9 & 51.3 & \phantom{0}+8.4\\
Cooking and Food                  & 47.9  & 47.7 & 52.7 & \phantom{0}+4.8 \\
Weather and Climate               & 51.7  & 46.3 & 54.8 & \phantom{0}+3.1\\
People and Everyday               & 43.1  & 44.4 & 51.5 & \phantom{0}+7.1\\
Objects, Material and Clothing    & 42.9  & 45.4 & 49.3 & \phantom{0}+3.9\\
Other                             & 41.5  & 50.2 & 51.2 & \phantom{0}+1.0 \\
\bottomrule
\end{tabular}}
\caption{Accuracy (\%) of question types in OK-VQA full testing set. Our models outperforms exp and imp models by a large margin on all categories. (exp: explicit-only model and imp: implicit-only model)}
\label{exp:category}
\vspace{-5mm}
\end{table}

\subsection{Qualitative Analysis}
\label{sec:qualitative}
Analyzed in previous sections, jointly reasoning over both knowledge sources during answer generation improves the explicit-only and implicit-only models by large margins. Figure~\ref{fig:examples} shows two examples comparing answers generated by different models along with retrieved knowledge. The left example shows that while explicit knowledge retrieved from the knowledge base contains the necessary knowledge entries for reasoning, it fails to generate the answer which requires the relation between bench and Coca Cola logos. On the other side, implicit knowledge retrieved from GPT-3 can only infer the bench is painted red, failing to recognize its logo. By jointly considering both knowledge sources, our model can associate the color of Coca Cola logo with the painted color of the bench which derives the correct answer. The right example shows that though explicit knowledge does not contain the right knowledge entries, it provides visually similar descriptions of this sport which further constrains the search space of our model and verifies the correctness of the implicit knowledge.

\section{Conclusion}
\label{sec:conclusion}
This paper takes a step towards understanding the complementary role of implicit knowledge gained from continuing to scale models and explicit knowledge from structured knowledge bases.  Importantly, it appears that there is headroom in both directions (i.g. improving retrieval and reasoning). Our conceptually simple yet effective approach for knowledge-based VQA makes these relationships explicit while still achieving a significant improvement against state-of-the-art results. 
Additional challenges remain, for example 
how best to align image regions with meaningful external semantics deserves and how to efficiently and accurately integrate multiple knowledge bases.

\section*{Acknowledgement}
We are especially grateful to Jianwei Yang, Daniel McDuff, Dragomir Radev, Harkirat Behl, Hao Chen, Chunyuan Li, Baolin Peng, Kezhen Chen, Tejas Srinivasan for their for the early insightful discussions, suggestion, and their pointers to the modeling generation and literature. We thank Zhe Gan, Zhengyuan Yang, Lijuan Wang from cognition service team of Microsoft for their work and their generous helps and feedback for the project. We appreciate Subhojit Som from Turing team of Microsoft for his enormous support and encouragement. The authors gratefully acknowledge Kenneth Marino from DeepMind, and Roozbeh Mottaghi from the AllenAI for their comments, supporting and helps of the work.

\bibliography{acl_latex}
\bibliographystyle{acl_natbib}

\clearpage

\begin{appendices}

\section*{Appendix}


\section{Figure of Explicit Knowledge}
\label{app:explicit}
\begin{figure}[h]
\centering
\includegraphics[width=\columnwidth]{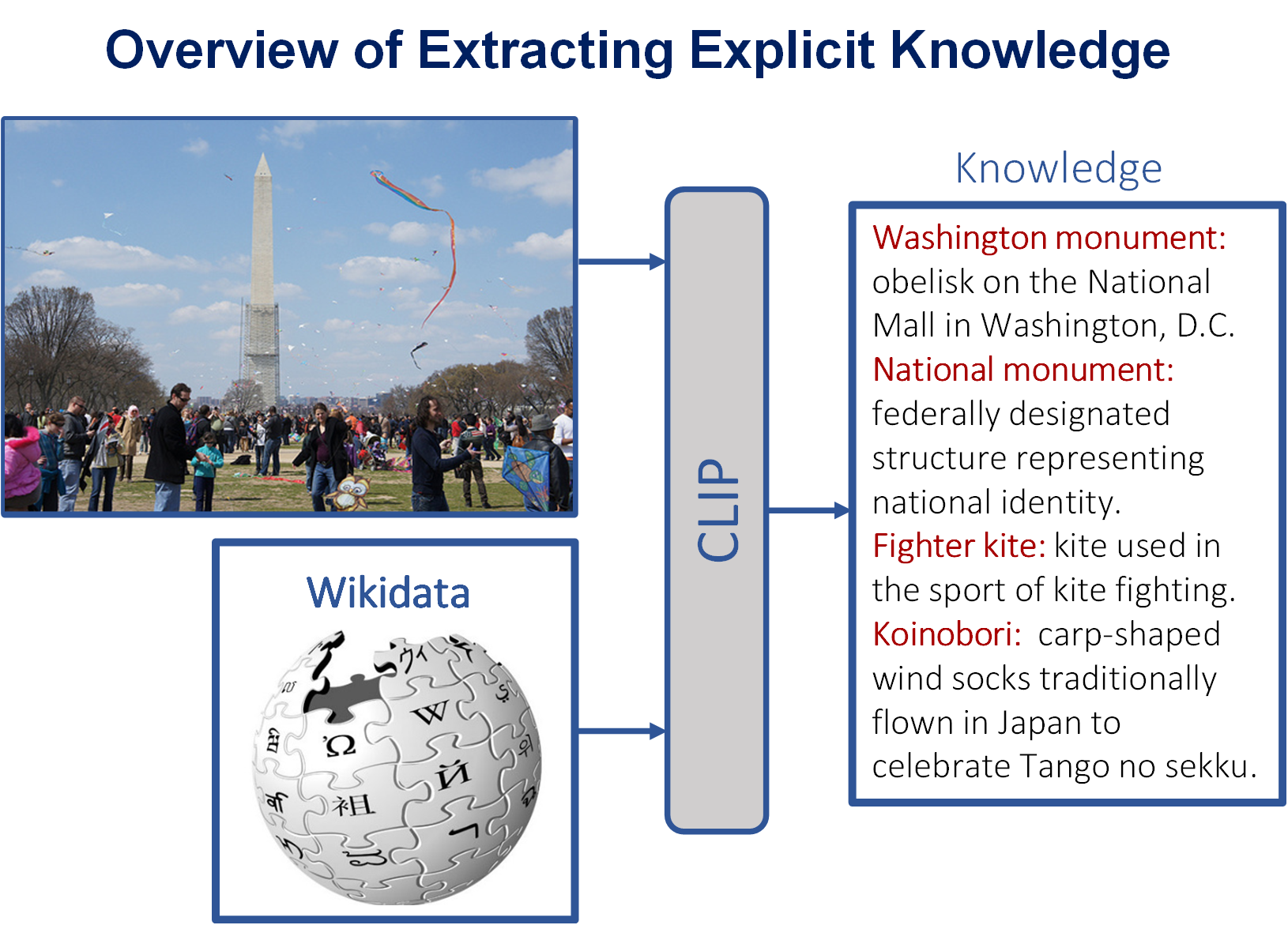}
\caption{Overview of the explicit knowledge extraction. We use a sliding window to crop image regions and retrieve knowledge entries from an explicit knowledge base by CLIP.}
\label{fig:knowledge-ext}
\vspace{-5mm}
\end{figure}

In this section, we show one example Figure~\ref{fig:knowledge-ext} to extract explicit knowledge from an image, which use the CLIP model to conduct the explicit knowledge retrieval with the image and a wiki knowledge base.

\section{Examples of Prompts of Implicit Knowledge}
\label{app:prompt}
\begin{figure}[h]
\centering
\includegraphics[width=.89\columnwidth]{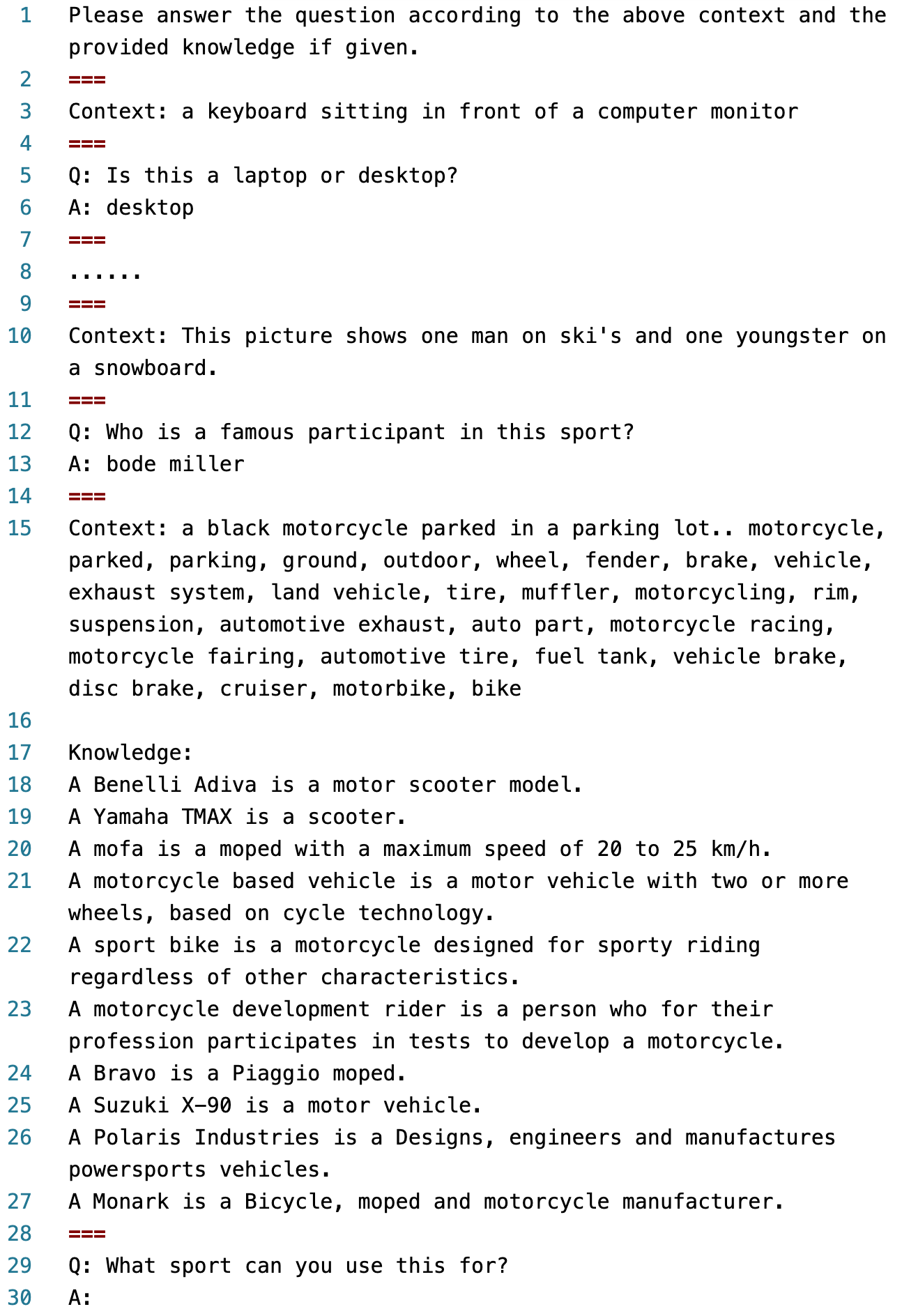}
\vspace{-3mm}
\caption{An example of the prompts that we use to query GPT-3 in our knowledge-aumented GPT-3 query system.
}
\label{fig:prompt2}
\vspace{-0.3cm}
\end{figure}

In this Section B of the Appendix, we show two concrete examples (Figure~\ref{fig:prompt1} and Figure~\ref{fig:prompt2}) for the prompts that constructed to query GPT-3 for implicit knowledge in our experiments:

\begin{figure}[h]
\centering
\includegraphics[width=.9\columnwidth]{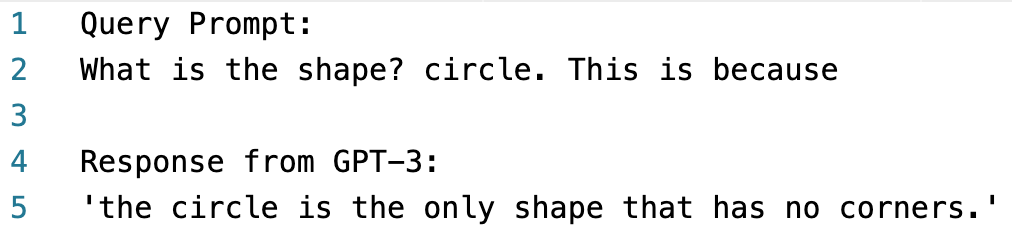}
\vspace{-3mm}
\caption{An example of the evidence of rationale that we obtain from GPT-3 by using a combination of question and answer candidate to query it.}
\label{fig:prompt1}
\vspace{-0.3cm}
\end{figure}

\section{Analysis on More Examples}
\label{app:analysis}
\begin{figure*}
\begin{minipage}[!ht]{\textwidth}
  \centering
  \includegraphics[trim={.8cm 6cm 1cm 0},clip, 
  width=0.75\textwidth]{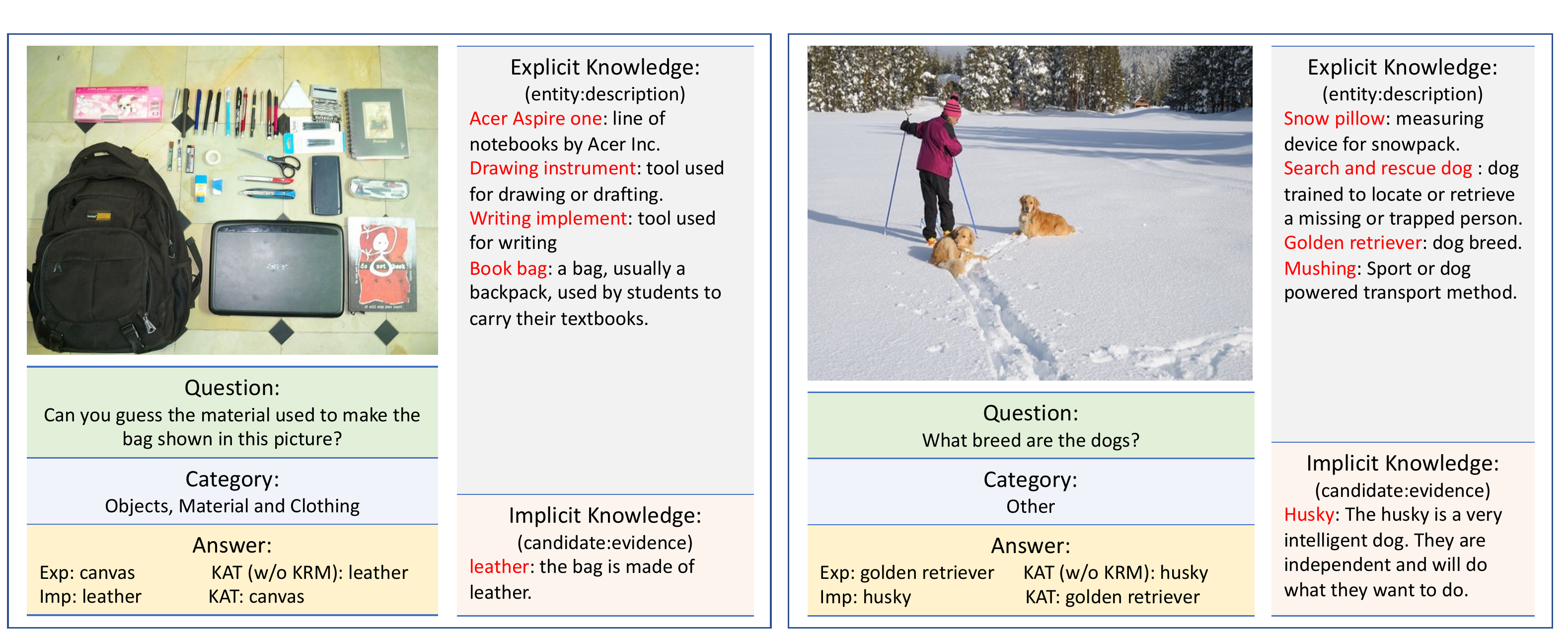}\\
        \includegraphics[trim={.8cm 6cm 1cm 0},clip, 
    width=0.75\textwidth]{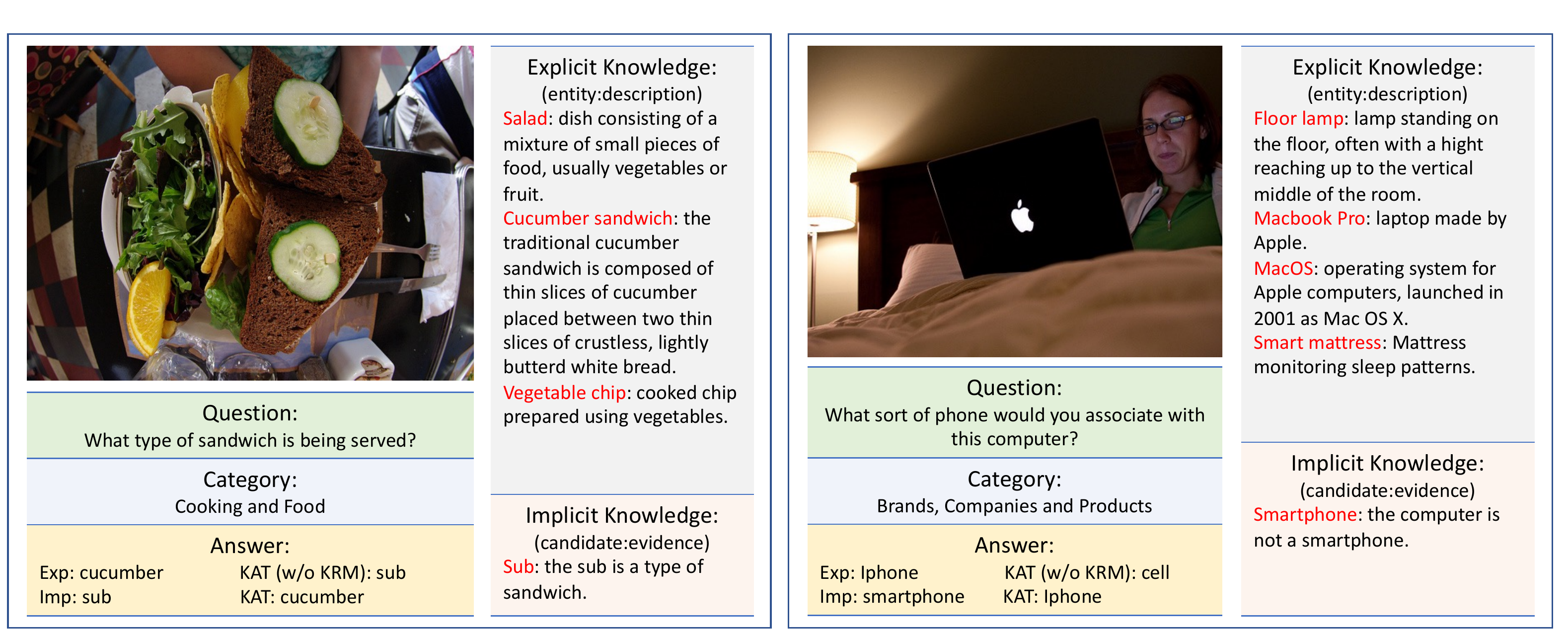}\\
        \includegraphics[trim={0.8cm 6cm 1cm 0},clip, 
    width=0.75\textwidth]{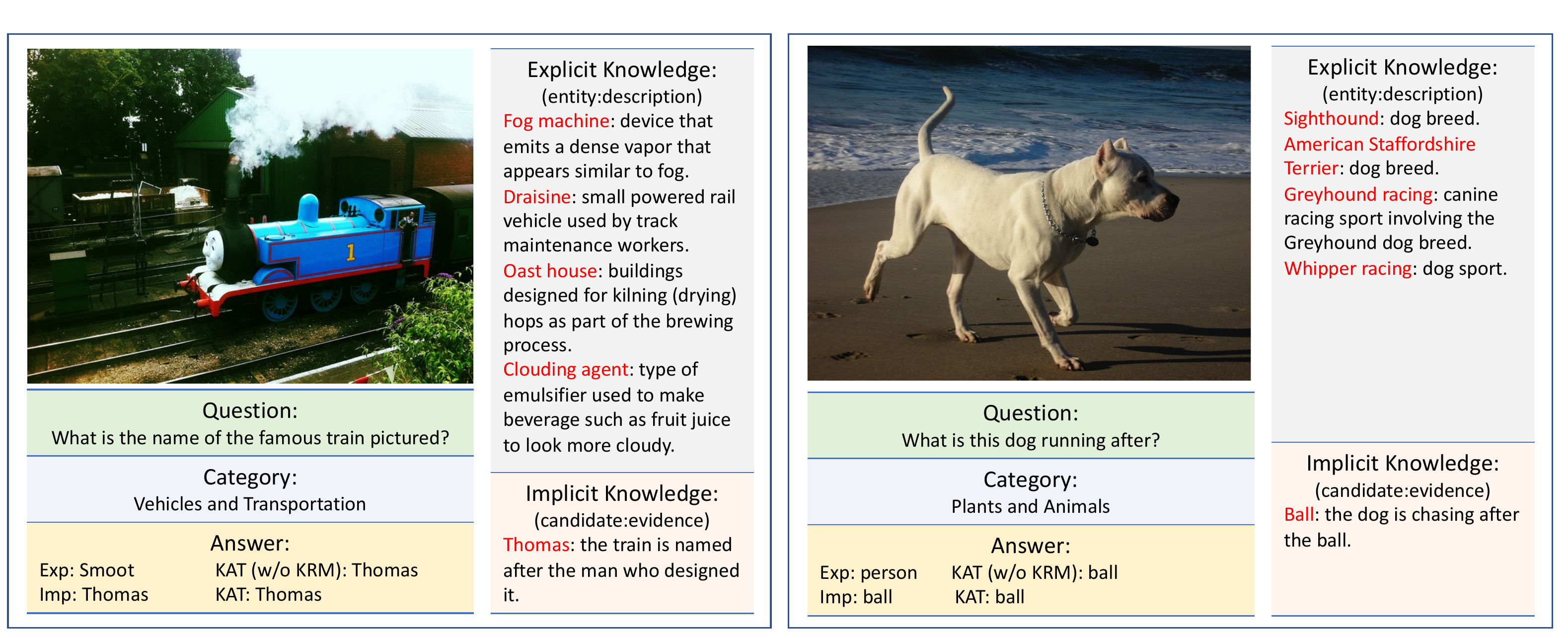}  \\
        \includegraphics[trim={0.8cm 6cm 1cm 0},clip, 
    width=0.75\textwidth]{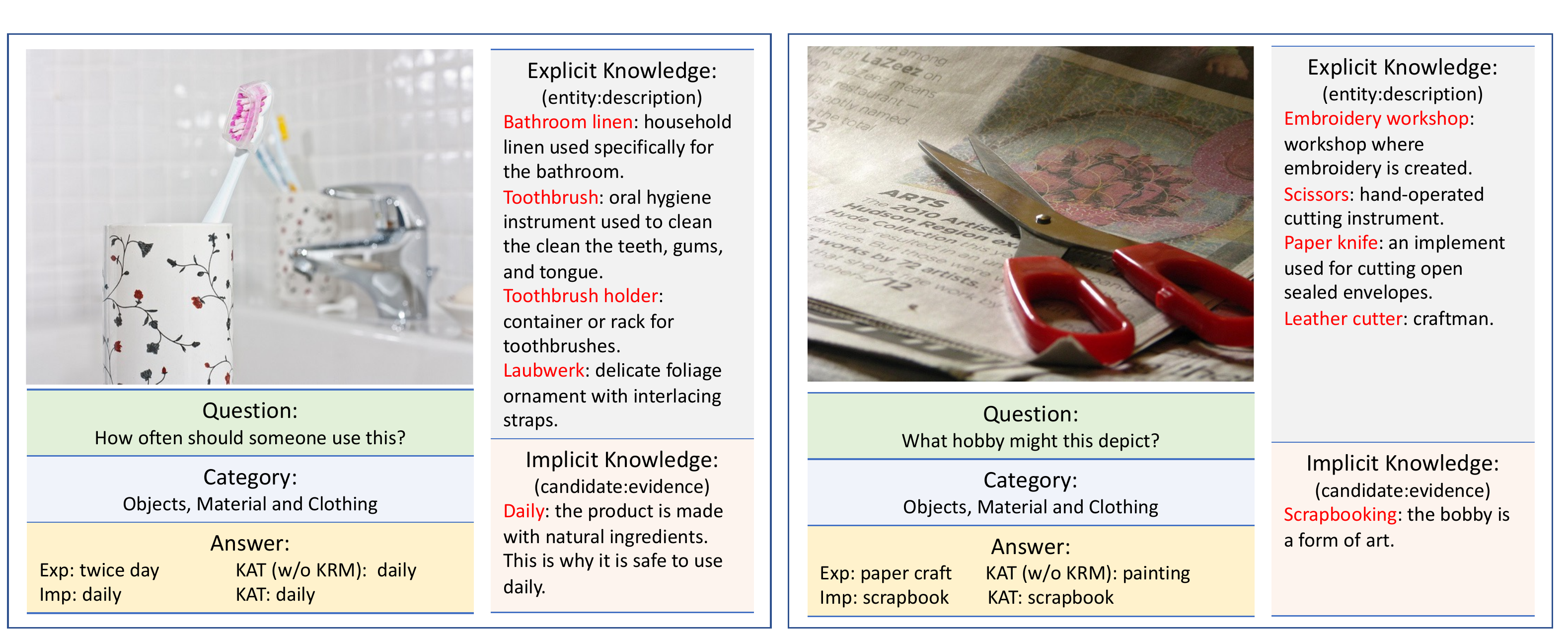} \\
        \includegraphics[trim={0.8cm 6cm 1cm 0},clip, 
    width=0.75\textwidth]{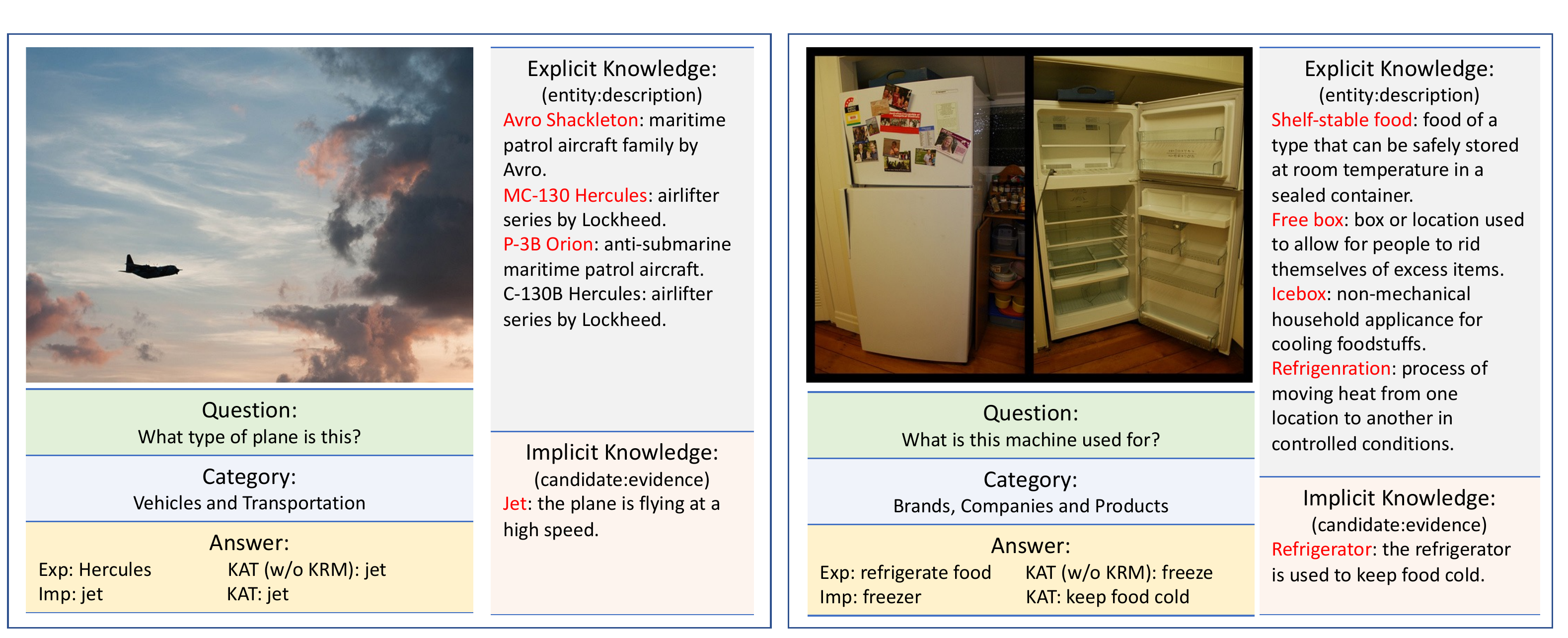}\\
\captionof{figure}{More examples from OK-VQA dataset that our model generates answers by jointly reasoning over both implicit and explicit knowledge.}
\label{fig:moreexamples}
\vspace{-3mm}
\end{minipage}
\end{figure*}

In this section, we showcase more predictions from variants of our model. As shown in Figure~\ref{fig:moreexamples}, we analyze the predictions based on different type of knowledge from several aspects:

\paragraph{Effectiveness of explicit knowledge retriever.} Our explicit knowledge retriever can retrieve fine-grained knowledge entries from the explicit knowledge base, such as \emph{golden retriever} (a fine-grained breed of dogs), \emph{cucumber sandwich} (a specific type of sandwich) and \emph{Macbook Pro} (a specific model of Apple products). These fine-grained entities are hardly obtained from existing object detection models, which can constraint the search space of our model and are beneficial to our answer generation process.

\paragraph{Effectiveness of implicit knowledge retriever.} Our implicit knowledge retriever can retrieve supporting evidence from GPT-3, such as \emph{Thomas: the train is named after the man who designed it.} and \emph{Refrigerator: the refrigerator is used to keep food cold}. These kinds of knowledge are highly related to commonsense knowledge which needs further inference based on entities and provide complementary explanation to explicit knowledge.

\paragraph{Answer generation \& classification.} As most previous work on OK-VQA task, such as KRISP or MAVEx method, implement OK-VQA as a classification task. The prediction vocabulary is dataset-specific and assumes the training and test set are sharing a similar vocabulary. The limitation of these methods is the generalization ability. Our proposed KAT model treats OK-VQA as an open-end generation task. From these examples we found, our model can generate answers like \emph{Iphone} or \emph{Hercules} that are visually and semantically reasonable. Our proposed novel KAT model using the explicit and implicit knowledge is designed to enhance semantic alignment and generate representations with stronger knowledge-awareness.

\end{appendices}
\end{document}